\pdfoutput=1
\documentclass[11pt]{article}

\usepackage[]{ACL2023}

\usepackage{times}
\usepackage{latexsym}

\usepackage[T1]{fontenc}

\usepackage[utf8]{inputenc}

\usepackage{microtype}

\usepackage{inconsolata}

\usepackage{graphicx}
\usepackage{booktabs}
\usepackage{amsmath} 
\usepackage{multirow}
\usepackage{amssymb} 
\usepackage{amsmath} 
\usepackage{array} 
\usepackage{makecell} 

\usepackage{color} 
\definecolor{premisecolor}{RGB}{253,255,49}
\definecolor{conclusioncolor}{RGB}{255,102,102}
\definecolor{negativecolor}{RGB}{138,130,255}
\definecolor{adversativecolor}{RGB}{53,254,255}
\definecolor{coordinatingcolor}{RGB}{0,204,0}

\title{IDOL: Indicator-oriented Logic Pre-training for Logical Reasoning}

\author{
Zihang Xu$^\dag$,
Ziqing Yang$^\dag$,
Yiming Cui$^\ddag$$^\dag$,
Shijin Wang$^\dag$$^\S$ \\
{$^\dag$State Key Laboratory of Cognitive Intelligence, iFLYTEK Research, China} \\
{$^\ddag$Research Center for SCIR, Harbin Institute of Technology, Harbin, China 
} \\
{$^\S$iFLYTEK AI Research (Central China), Wuhan, China
} \\
$^\dag$\tt\{zhxu13,zqyang5,ymcui,sjwang3\}@iflytek.com \\
$^\ddag$\tt ymcui@ir.hit.edu.cn}

\begin{document}
\maketitle

\begin{abstract}


In the field of machine reading comprehension (MRC), existing systems have surpassed the average performance of human beings in many tasks like SQuAD. However, there is still a long way to go when it comes to logical reasoning. Although some methods for it have been put forward, they either are designed in a quite complicated way or rely too much on external structures. In this paper, we proposed \textbf{IDOL} (\textbf{I}n\textbf{D}icator-\textbf{O}riented \textbf{L}ogic Pre-training), an easy-to-understand but highly effective further pre-training task which logically strengthens the pre-trained models with the help of 6 types of logical indicators and a logically rich dataset \textbf{LGP} (\textbf{L}o\textbf{G}ic \textbf{P}re-training). IDOL achieves state-of-the-art performance on ReClor and LogiQA, the two most representative benchmarks in logical reasoning MRC, and is proven to be capable of generalizing to different pre-trained models and other types of MRC benchmarks like RACE and SQuAD 2.0 while keeping competitive general language understanding ability through testing on tasks in GLUE. Besides, at the beginning of the era of large language models, we take several of them like ChatGPT into comparison and find that IDOL still shows its advantage.\footnote{Please refer to \url{https://github.com/GeekDream-x/IDOL}  for relevant resources including datasets, models, and codes.}

\end{abstract}

\section{Introduction}

\begin{figure*}
\centering
\includegraphics[scale=0.283]{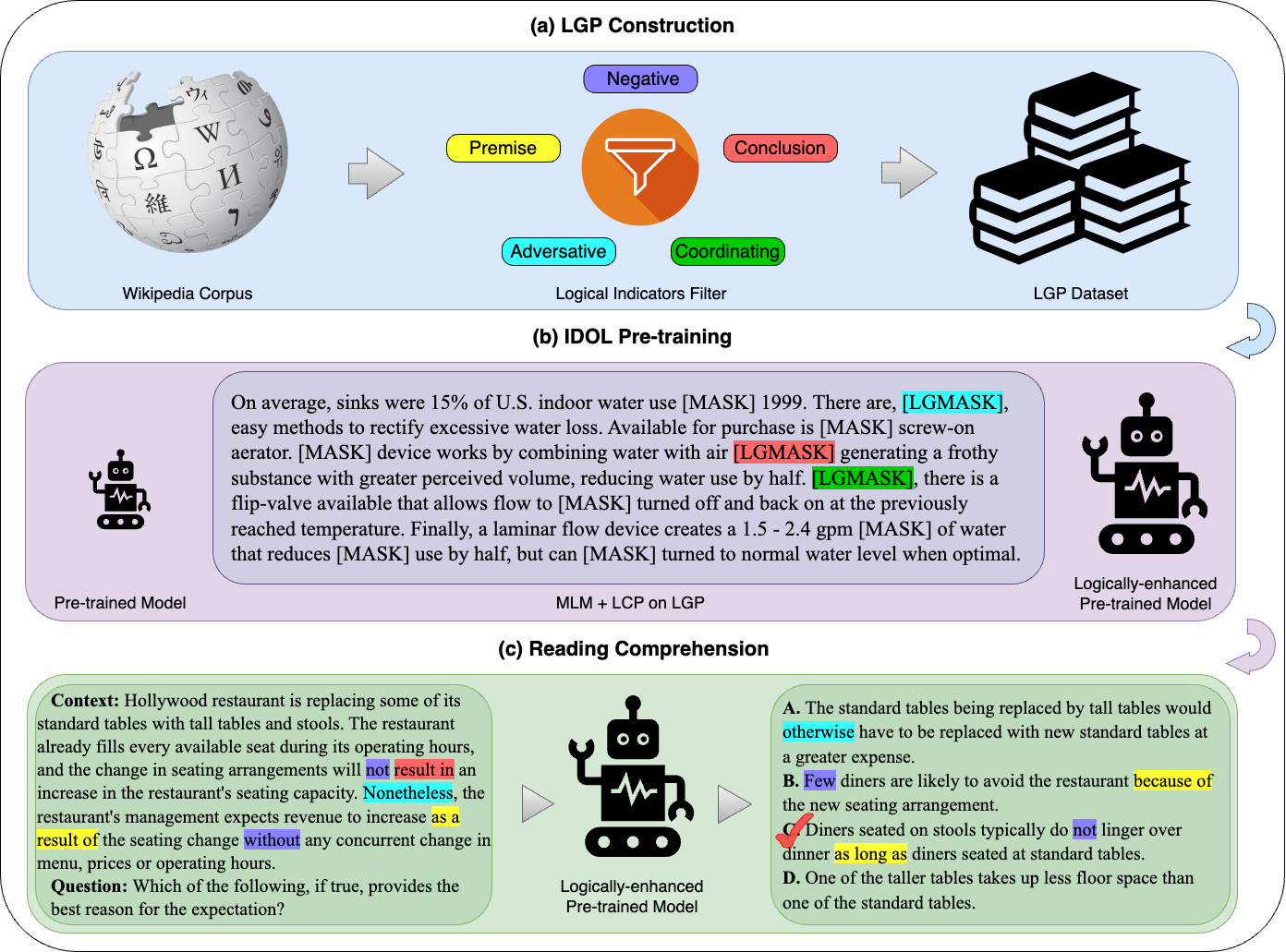}
\caption{A diagram illustrating the three steps of our method: (a) construct the logically rich dataset LGP from Wikipedia, (b) further pre-train models to improve logical reasoning ability, and (c) answer logical reasoning MRC questions with the help of logical indicators appeared both in context and choices. See Section \ref{method} for more details on our method.}
\label{fig:systemflowchart}
\end{figure*}

With the development of pre-trained language models, a large number of tasks in the field of natural language understanding have been dealt with quite well. However, those tasks emphasize more on assessing basic abilities like word-pattern recognition of the models while caring less about advanced abilities like reasoning over texts \cite{reasonanalysis}. 

In recent years, an increasing number of challenging tasks have been brought forward gradually. At sentence-level reasoning, there is a great variety of benchmarks for natural language inference like QNLI \cite{qnli} and MNLI \cite{mnli}. Although the construction processes are different, nearly all these datasets evaluate models with binary or three-way classification tasks which need reasoning based on two sentences. At passage-level reasoning, the most difficult benchmarks are generally recognized as the ones related to logical reasoning MRC which requires question-answering systems to fully understand the whole passage, extract information related to the question and reason among different text spans to generate new conclusions in the logical aspect. In this area, the most representative benchmarks are some machine reading comprehension datasets like ReClor \cite{reclor} and LogiQA \cite{logiqa}.


 Considering that there are quite few optimization strategies for the pre-training stage and that it is difficult for other researchers to follow and extend the existing methods which are designed in rather complex ways, we propose an easy-to-understand but highly effective pre-training task named IDOL which helps to strengthen the pre-trained models in terms of logical reasoning. We apply it with our customized dataset LGP which is full of logical information. Moreover, we experimented with various pre-trained models and plenty of different downstream tasks and proved that IDOL is competitive while keeping models and tasks agnostic.

Recently, ChatGPT attracts a lot of attention all over the world due to its amazing performance in question answering. Thus, we also arranged an experiment to let IDOL compete with a series of LLMs (large language models) including it.

The contributions of this paper are summarized as follows:
\begin{itemize}
    \item Put forward the definitions of 5 different types of logical indicators. Based on these we construct the dataset LGP for logical pre-training and we probe the impact of different types of logical indicators through a series of ablation experiments.
    \item Design an indicator-oriented further pre-training method named IDOL, which aims to enhance the logical reasoning ability of pre-trained models. It achieves state-of-the-art performance in logical reasoning MRC and shows progress in general MRC and general understanding ability evaluation.
    \item The first to provide a pilot test about the comparison between fine-tuning traditional pre-trained models and prompting LLMs in the field of logical reasoning MRC.

\end{itemize}

\begin{table*}
\small
\begin{tabular}{m{0.6cm}<{\centering}p{8cm}p{6cm}}
\toprule[1pt]
\textbf{Type} & \textbf{Library} & \textbf{Example}  \\ \midrule
\textbf{PMI} & given that, seeing that, for the reason that, owing to, as indicated by, on the grounds that, on account of, considering, because of, due to, now that, may be inferred from, by virtue of, in view of, for the sake of, thanks to, as long as, based on that, as a result of, considering that, inasmuch as, if and only if, according to, in that, only if, because, depend on, rely on & The real world contains no political entity exercising literally total control over even one such aspect. \colorbox{premisecolor}{This is because} any system of control is inefficient, and, therefore, its degree of control is partial. \\ \midrule

\textbf{CLI} & conclude that, entail that, infer that, that is why, therefore, thereby, wherefore, accordingly, hence, thus, consequently, whence, so that, it follows that, imply that, as a result, suggest that, prove that, as a conclusion, conclusively, for this reason, as a consequence, on that account, in conclusion, to that end, because of this, that being so, ergo, in this way, in this manner, by such means, as it turns out, result in, in order that, show that, eventually & In the United States, each bushel of corn produced might \colorbox{conclusioncolor}{result in} the loss of as much as two bushels of topsoil. Moreover, in the last 100 years, the topsoil in many states, which once was about fourteen inches thick, has been eroded to only six or eight inches. \\ \midrule

\textbf{NTI} & not, neither, none of, unable, few, little, hardly, merely, seldom, without, never, nobody, nothing, nowhere, rarely, scarcely, barely, no longer, isn't, aren't, wasn't, weren't, can't, cannot, couldn't, won't, wouldn't, don't, doesn't, didn't, haven't, hasn't & A high degree of creativity and a high level of artistic skill are \colorbox{negativecolor}{seldom} combined in the creation of a work of art. \\ \midrule

\textbf{ATI} & although, though, but, nevertheless, however, instead of, nonetheless, yet, rather, whereas, otherwise, conversely, on the contrary, even, nevertheless, despite, in spite of, in contrast, even if, even though, unless, regardless of, reckless of & This advantage accruing to the sentinel does not mean that its watchful behavior is entirely self-interested. \colorbox{adversativecolor}{On the contrary}, the sentinel's behavior is an example of animal behavior motivated at least in part by altruism. \\ \midrule

\textbf{CNI} & and, or, nor, also, moreover, in addition, on the other hand, meanwhile, further, afterward, next, besides, additionally, meantime, furthermore, as well, simultaneously, either, both, similarly, likewise & A graduate degree in policymaking is necessary to serve in the presidential cabinet. \colorbox{coordinatingcolor}{In addition}, everyone in the cabinet must pass a security clearance. \\ 

\bottomrule[1pt]
\end{tabular}
\caption{Libraries and examples of all types of logical indicators.}
\label{table:lgidclibandexp}
\end{table*}

\section{Related Work}

\subsection{Logical Reasoning}
In order to help reasoning systems perform better on reading comprehension tasks focusing on logical reasoning, there have been a great many methods put forward by research institutions from all over the world. Unsurprisingly, the majority of the optimization approaches put forward revolve around the fine-tuning phase while there are far fewer methods designed for further pre-training. 

In the aspect of pre-training, to the best of our knowledge, there are only two approaches presented in published papers called MERIt and LogiGAN. MERIt team generated a dataset from the one provided by \citet{meritdataorigin} which contains passages from Wikipedia with annotations about entities and relations. And then optimize the model on that with the help of contrastive learning \cite{jiao-etal-2022-merit}. The researchers behind LogiGAN use a task about statement recovery to enhance the logic understanding ability of generative pre-trained language models like T5 \cite{logigan}.

For optimizing models at the fine-tuning phase, there are dozens of methods proposed as far as we know. For example, LReasoner put forward a context extension framework with the help of logical equivalence laws including contraposition and transitive laws \cite{lreasoner}. Another example is Logiformer which introduced a two-stream architecture containing a syntax branch and a logical branch to better model the relationships among distant logical units \cite{logiformer}.

\subsection{Pre-training Tasks}

As NLP enters the era of pre-training, more and more researchers are diving into the design of pre-training tasks, especially about different masking strategies. For instance, in \citet{wwm}, the authors apply Whole Word Masking (WWM) on Chinese BERT and achieved great progress. WWM changes the masking strategy in the original masked language modeling (MLM) into masking all the tokens which constitute a word with complete meaning instead of just one single token.
In addition, \citet{tlm} extends MLM to parallel data as Translation Language Modeling (TLM) which randomly masks tokens in both source and target sentences in different languages simultaneously. The results show that TLM is beneficial to improve the alignment among different languages.

\section{Preliminary}

\subsection{Text Logical Unit}



It is admitted that a single word is the most basic unit of a piece of text but its meaning varies with different contexts. In \citet{logiformer}, the authors refer logical units to the split sentence spans that contain independent and complete semantics. In this paper, since much more abundant logical indicators with different types that link not only clauses but also more fine-grained text spans are introduced, we extend this definition to those shorter text pieces like entities.

\subsection{Logical Indicators} \label{logicindicators}

By analyzing the passages in logical reasoning MRC and reasoning-related materials like debate scripts, we found that the relations between logic units (like entities or events) can be summarized into 5 main categories as follows and all these relations are usually expressed via a series of logical indicators. After consulting some previous work like \citet{logigan} and Penn Discourse TreeBank 2.0 (PDTB 2.0) \cite{pdtb2}, we managed to construct an indicator library for each category. As for the examples of indicators we used in detail, please refer to Table \ref{table:lgidclibandexp}. 

\begin{itemize}
    \item \textbf{Premise/Conclusion Indicator (PMI/CLI)} The first two types of logical indicators pertain to premises and conclusions. These indicators signal the logical relationship between statements. For instance, premise expressions such as ``due to'' indicate that the logic unit following the keyword serves as the reason or explanation for the unit preceding it. Conversely, conclusion phrases like ``result in'' suggest an inverse relationship, implying that the logic unit after the keyword is a consequence or outcome of the preceding unit.
    \item \textbf{Negative Indicator (NTI)} Negative indicators, such as ``no longer'', play a crucial role in text logic by negating affirmative logic units. They have the power to significantly alter the meaning of a statement. For example, consider the sentences ``Tom likes hamburgers.'' and ``Tom no longer likes hamburgers.'' These two sentences have nearly opposite meanings, solely due to the presence of the indicator ``no longer''.
    \item \textbf{Adversative Indicator (ATI)} Certain expressions, such as ``however'', are commonly employed between sentences to signify a shift or change in the narrative. They serve as valuable tools for indicating the alteration or consequence of a preceding event, which helps to cover this frequent kind of relation among logic units.
    \item \textbf{Coordinating Indicator (CNI)} The coordinating relation is undoubtedly the most prevalent type of relationship between any two logic units. Coordinating indicators are used to convey that the units surrounding them possess the same logical status or hold equal importance. These indicators effectively demonstrate the coordination or parallelism between the connected logic units.

\end{itemize}

\section{Methodology} \label{method}

\subsection{LGP Dataset Construction} \label{lgpdatasetconstruction}

For the sake of further pre-training models with IDOL, we constructed the dataset LGP (LoGic Pre-training) based on the most popular unannotated corpus English Wikipedia.\footnote{\url{https://dumps.wikimedia.org/}} We first split the articles into paragraphs and abandoned those whose lengths (after tokenization) were no longer than 5. In order to provide as much logical information as possible, we used the logical indicators listed in Table \ref{table:lgidclibandexp} to filter the Wiki paragraphs. During this procedure, we temporarily removed those indicators with extremely high frequency like ``and'', otherwise, there would be too many paragraphs whose logical density was unacceptably low. Then, we iterated every logical keyword and replaced it with our customized special token \texttt{[LGMASK]} under the probability of 70\%. 

\begin{figure}
\centering
\includegraphics[scale=0.15]{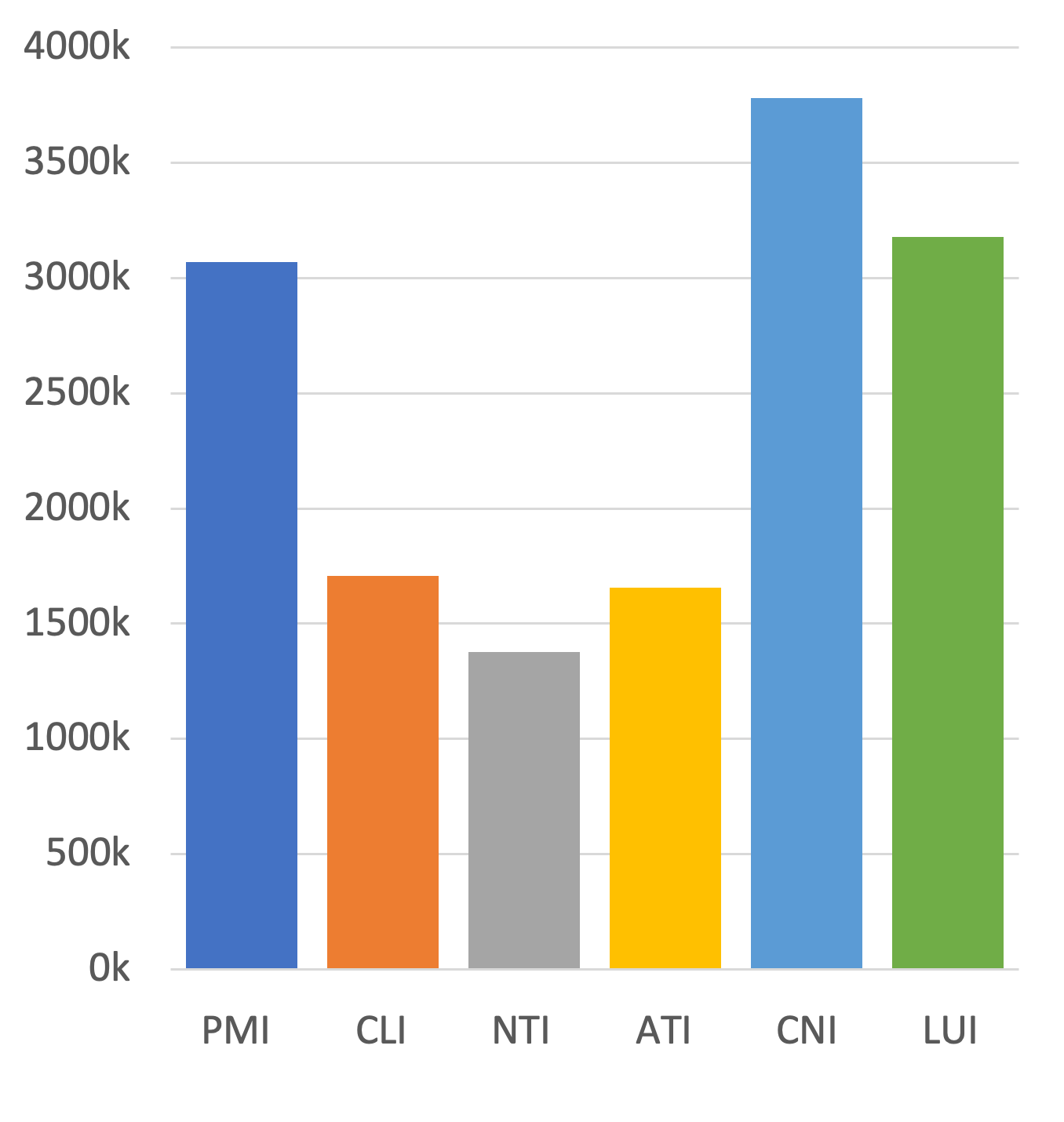}
\caption{The numbers of 6 types of logical indicators in LGP for RoBERTa.}
\label{fig:lgplgmasknum}
\end{figure}

For the purpose of modeling the ability to distinguish whether a certain masked place is logic-related or not, we introduced the sixth logical indicator type - Logic Unrelated Indicator (LUI). Based on this, we then randomly replaced 0.6\% tokens other than logical indicators with \texttt{[LGMASK]}. Afterward, the labels for the logical category prediction (LCP) task were generated based on the corresponding logic types of all the \texttt{[LGMASK]}s. In the end, take RoBERTa \cite{roberta} for example, our logic dataset LGP contains over 6.1 million samples and as for the quantities of logical indicators in each type please refer to Figure~\ref{fig:lgplgmasknum}.

\subsection{IDOL Pre-training}

\subsubsection{Logical Category Prediction} 

As introduced in section \ref{logicindicators} and section \ref{lgpdatasetconstruction}, we defined a logic-related special mask token \texttt{[LGMASK]} and it will take the place of 6 types of logical indicators - PMI, CLI, NTI, ATI, CNI, and LUI. During the forward process of fine-tuning the pre-trained models, the corresponding logical categories need to be predicted by them like what will be done in the token classification task of the standard Masked Language Modeling (MLM) \cite{bert}.

When the models are trying to predict the correct logical type of a certain \texttt{[LGMASK]}, they will learn to analyze the relationship among the logical units around the current special token and whether there is some kind of logical relations with the help of the whole context. Therefore, the pre-trained models will be equipped with a stronger ability of reasoning over texts gradually. 

Moreover, we use Cross-Entropy Loss (CELoss) to evaluate the performance of predicting the logical categories. The loss function for LCP is as described in Equation \eqref{lcplossfunction} where $ n $ is the number of samples, $ m $ is the number of \texttt{[LGMASK]} in the $ i_{th} $ sample, $ y_{i,j} $ indicates the model prediction result for the $ j_{th} $ \texttt{[LGMASK]} in the $ i_{th} $ sample and $ \hat{y}_{i,j} $ denote the corresponding ground truth value.

\begin{equation}
\begin{aligned}
\mathcal{L}_{\textrm{LCP}} = \sum_{i=1}^{n} \frac{1}{m}\sum_{j=1}^{m} \textrm{CELoss}(y_{i,j},\hat{y}_{i,j}) \label{lcplossfunction}
\end{aligned}
\end{equation}

\subsubsection{IDOL} 
\begin{figure}
\centering
\includegraphics[scale=0.35]{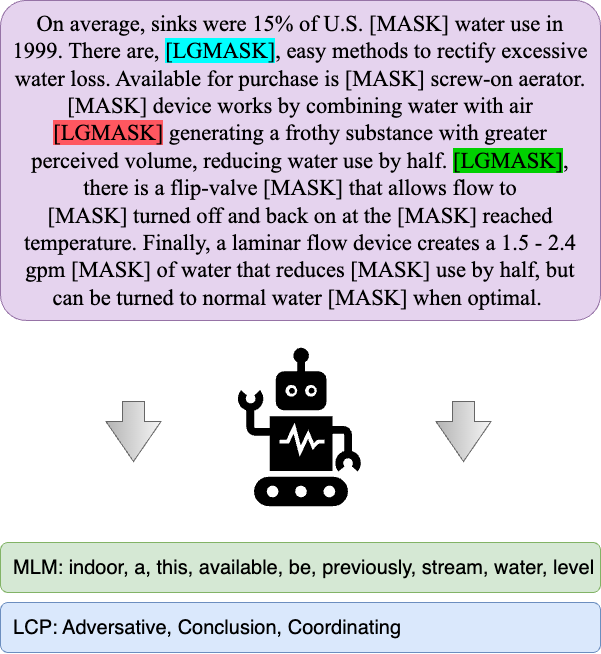}
\caption{An example of pre-training with IDOL. The model needs to recover the tokens replaced by [MASK] (MLM) and predict the category of each logical indicator masked by \texttt{[LGMASK]} (LCP) in the meantime.}
\label{fig:idolcase}
\end{figure}

To avoid catastrophic forgetting, we combine the classic MLM task with the LCP introduced above to become IDOL, a multi-task learning pre-training method for enhancing the logical reasoning ability of pre-trained models. For the purpose of balancing the effects of the two pre-training tasks, we introduced a hyper-parameter $\lambda$ as the weight of the loss of LCP (the proper $\lambda$ depends on the pre-trained language model used and the empirical range is between 0.7 and 0.9). Thus, for the IDOL pre-training loss function, please refer to Equation \eqref{idollossfunction}. Figure \ref{fig:idolcase} presented an example of IDOL pre-training where predicting tokens and the classes of logical indicators simultaneously.

\begin{equation}
\begin{aligned}
\mathcal{L}_{\textrm{IDOL}} = \lambda \cdot	\mathcal{L}_{\textrm{LCP}} + (1-\lambda) \cdot\mathcal{L}_{\textrm{MLM}} \label{idollossfunction}
\end{aligned}
\end{equation}

\section{Experiments}

\subsection{Baselines}

With the rapid development of pre-training technology these years, we have various choices for backbone models. In this paper, we decide to apply IDOL on BERT-large \cite{bert}, RoBERTa-large \cite{roberta}, ALBERT-xxlarge \cite{albert} and DeBERTa-v2-xxlarge \cite{deberta} and will evaluate the models in the following three different aspects in section \ref{mainresults} to better verify the performance of IDOL.\footnote{In the following sections, we refer these baseline models to BERT, RoBERTa, ALBERT and DeBERTa respectively for simplicity.} 

In terms of logical reasoning MRC, we will compare IDOL with several previous but still competitive methods for logical reasoning MRC including DAGN \cite{huang-etal-2021-dagn}, AdaLoGN \cite{li-etal-2022-adalogn}, LReasoner \cite{wang-etal-2022-logic}, Logiformer \cite{logiformer} and MERIt \cite{jiao-etal-2022-merit}. Much more interesting, we let IDOL compete with ChatGPT in a small setting.

\subsection{Datasets}

First and foremost, the aim of IDOL is to improve the logical reasoning ability of pre-trained models, thus, the two most representative benchmarks - ReClor and LogiQA will act as the primary examiners.

Following this, RACE \cite{lai-etal-2017-race} and SQuAD 2.0 \cite{squad2.0}, two classic machine reading comprehension datasets that are not targeted at assessing reasoning ability, will come on stage, which will be beneficial to conclude whether IDOL helps with other types of reading comprehension abilities.

Last but not least, we also tested the models pre-trained with IDOL on MNLI \cite{mnli} and STS-B \cite{stsb}, two tasks of GLUE \cite{glue}, to make sure that the general language understanding abilities are retained to a great extent during the process of logical enhancement. The evaluation metrics on STS-B are the Pearson correlation coefficient (Pear.) and Spearman's rank correlation coefficient (Spear.) on the development set. And we use the accuracy of MNLI-m and MNLI-mm development sets for evaluation on MNLI.

\textbf{ReClor} The problems in this dataset are collected from two American standardized tests - LSAT and GMAT, which guarantee the difficulty of answering the questions. Moreover, ReClor covers 17 classes of logical reasoning including main idea inference, reasoning flaws detection, sufficient but unnecessary conditions, and so forth. Each problem consists of a passage, a question, and four answer candidates, like the one shown in the green section of Figure \ref{fig:systemflowchart}. There are 4638, 500, and 1000 data points in the training set, development set, and test set respectively. The accuracy is used to evaluate the system's performance.

\textbf{LogiQA} The main difference compared with ReClor is that the problems in LogiQA are generated based on the National Civil Servants Examination of China. Besides, it incorporates 5 main reasoning types such as categorical reasoning and disjunctive reasoning. And 7376, 651, and 651 samples are gathered for the training set, development set, and test set individually.

\begin{table}[t]
\centering
\begin{tabular}{lcccc}
\toprule
\multirow{2}*{\textbf{Models}} & \multicolumn{2}{c}{\textbf{ReClor}} & \multicolumn{2}{c}{\textbf{LogiQA}} \\
 & Dev & Test & Dev & Test \\ \midrule
\textbf{BERT} & 53.8 & 49.8 & 35.3\textsuperscript{$ \spadesuit $} & 33.0\textsuperscript{$ \spadesuit $} \\ 
IDOL & \textbf{56.8} & \textbf{53.3} & \textbf{36.9} & \textbf{34.3} \\ \midrule 
\textbf{RoBERTa} & 62.6 & 55.6 & 37.0\textsuperscript{$ \spadesuit $} & 36.6\textsuperscript{$ \spadesuit $} \\  
DAGN & 65.2 & 58.2 & 35.5 & 38.7 \\ 
AdaLoGN & 65.2 & 60.2 & 39.9 & 40.7 \\ 
LReasoner & 66.2 & 62.4 & 38.1 & 40.6 \\   
MERIt & 67.8 & 60.7 & 42.4 & 41.5 \\
Logiformer & 68.4 & 63.5 & 42.2 & \textbf{42.6} \\  
IDOL & \textbf{70.2} & \textbf{63.9} & \textbf{42.5} & 41.8 \\ \midrule 
\textbf{ALBERT} & 70.4 & 67.3 & 41.2\textsuperscript{$ \spadesuit $} & 41.3\textsuperscript{$ \spadesuit $} \\ 
LReasoner & 73.2 & 70.7 & 41.6 & 41.2 \\
MERIT & 73.2 & \textbf{71.1} & 43.9 & \textbf{45.3} \\ 
IDOL & \textbf{74.6} & 70.9 & \textbf{44.7} & 43.8 \\  

\bottomrule
\end{tabular}
\caption{\label{rlmainresults}Results on logical reasoning MRC benchmarks - ReClor and LogiQA. In each block, the previous methods listed for comparison and IDOL take the pre-trained model in the first line as their backbone model. $ \spadesuit $: reproduced by ourselves.} 
\end{table}

\begin{table}[t]
\centering
\begin{tabular}{lccc}
\toprule
\multirow{2}*{\textbf{Models}} & \multicolumn{3}{c}{\textbf{ReClor}} \\ 
 & Test & Test-E & Test-H \\ \midrule
DeBERTa\textsuperscript{$ \heartsuit $} & 75.3 & 84.0 & 68.4 \\ 
LReasoner\textsuperscript{$ \clubsuit $}  & 76.1 & 87.1 & 67.5 \\ 
Knowledge Model\textsuperscript{$ \clubsuit $} & 79.2 & \textbf{91.8} & 69.3 \\ 
MERIt\textsuperscript{$ \clubsuit $} & 79.3 & 85.2 & 74.6 \\ 
AMR-LE\textsuperscript{$ \clubsuit $} & 80.0 & 87.7 & 73.9 \\
IDOL & \textbf{80.6} & 87.7 & \textbf{75.0} \\ 
\bottomrule
\end{tabular}
\caption{\label{reclordeberta}Results of IDOL with DeBERTa and other publicly available data. $ \clubsuit $: top results from the official leaderboard of ReClor (as of January 19, 2023). $ \heartsuit $: the performance of the original DeBERTa from \citet{jiao-etal-2022-merit} for reference (the majority of the top submissions and IDOLs in this table take DeBERTa as the backbone model).}
\end{table}


\subsection{Implementation Detail} \label{implementationdetail}

\subsubsection{IDOL}
During the process of pre-training with IDOL, we implemented the experiments on 8 Nvidia A100 GPUs. Since IDOL was applied on multiple different pre-trained models, we provide a range for some main hyperparameters. The whole training process consists of 10k\textasciitilde20k steps while the warm-up rate keeps 0.1. The learning rate is warmed up to a peak value between 5e-6\textasciitilde3e-5 for different models, and then linearly decayed. As for batch size, we found that 1024 or 2048 is more appropriate for most models. Additionally, we use AdamW \cite{adamw} as our optimizer with a weight decay of around 1e-3. For the software packages we used in detail, please see Appendix. 

With respect to the hyperparameters for fine-tuning models on downstream tasks, we follow the configurations provided in the original paper of either the corresponding model or the dataset.

\subsubsection{LLM} \label{llmimpledetail}
For the purpose of comparing IDOL with LLMs, we randomly sampled 30 pieces of data in the development sets of ReClor and LogiQA separately (named Dev-30). As for models, we choose GPT-3.5\footnote{The exact version is text-davinci-003.}, ChatGPT\footnote{Tested on February 13th, 2023.} and GLM-130B \cite{glm130b} for this pilot test.

To better evaluate the performance of LLMs, we tested them in the following three settings: zero-shot prompting, few-shot prompting, and chain-of-thought prompting. For zero-shot prompting, we designed the following template to wrap up the MRC problem.

\begin{quote}\small
\textit{The passage is \texttt{[PASSAGE]}. The question is \texttt{[QUESTION]}. Here are 4 choices for it and they are \texttt{[CHOICES]}. Which one should I choose? Thanks.} 
\end{quote}

\begin{figure}
\centering
\includegraphics[scale=0.15]{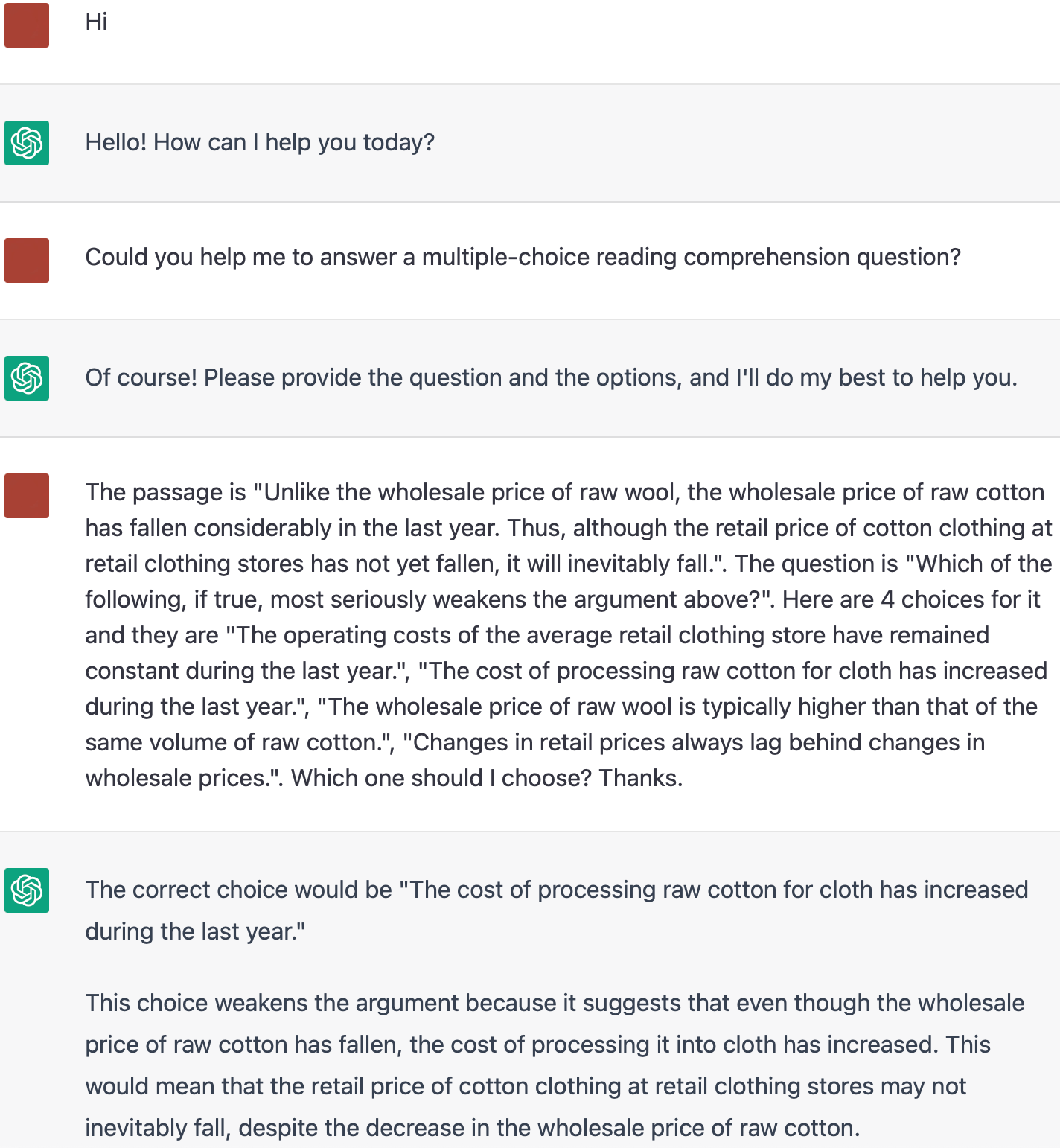}
\caption{An example of ChatGPT answering an MRC question.}
\label{fig:chatgptchathistory}
\end{figure}

As for few-shot prompting, we insert 3 examples in the same template but with correct answers ahead of the target question. When testing with chain-of-thought prompting, the template is similar to the one presented above. But there is only one example ahead and sentences describing the process of the way how humans reason to solve the problem are provided before giving the right answer to the example. For more details about the templates and the test example, please refer to Table \ref{table:llmtemplates} and Figure \ref{fig:chatgptchathistory}.

\begin{table*}[t]
\centering
\setlength\tabcolsep{5pt} 
\begin{tabular}{lcccccccccccc}
\toprule
\multirow{2}*{\textbf{Models}} & \multicolumn{2}{c}{\textbf{ReClor}} & \multicolumn{2}{c}{\textbf{LogiQA}} & \multicolumn{2}{c}{\textbf{RACE}} & \multicolumn{2}{c}{\textbf{SQuAD 2.0}} & \multicolumn{2}{c}{\textbf{STS-B}} & \multicolumn{2}{c}{\textbf{MNLI}} \\ 
 & Dev & Test & Dev & Test & Dev & Test & F1 & EM & Pear. & Spear. & m & mm \\ \midrule
RoBERTa & 62.7 & 55.2 & 36.2 & 37.1 & 85.2 & 84.4 & 89.0 & 86.1 & \textbf{92.6} & \textbf{92.5} & 89.5 & 89.3 \\
\hspace*{3mm} +MLM & 65.0 & 58.4 & 37.9 & 36.6 & 85.4 & 84.5 & 89.0 & 86.1& 92.2 & 92.1 & 89.5 & 89.5  \\   
\hspace*{6mm} +LCP (IDOL) & \textbf{66.8} & \textbf{60.6} & \textbf{39.4} & \textbf{38.8} & \textbf{85.6} & \textbf{84.8} & \textbf{89.2} & \textbf{86.2} & 92.3 & 92.2 & \textbf{89.7} & \textbf{89.5} \\ 
\bottomrule
\end{tabular}
\caption{\label{robertageneralizationtable}Results of RoBERTa with different pre-training tasks on logical reasoning MRC, other types of MRC and other types of NLU tasks.}
\end{table*}

\subsection{Main Results} \label{mainresults}


\subsubsection{Logical Reasoning MRC} 

\textbf{Fine-tuning} To evaluate the model performance on logical reasoning MRC, we experimented with the baseline models mentioned above on ReClor and LogiQA, the two most representative benchmarks in this field. The majority of previous researchers focus on applying their method to RoBERTa, IDOL meets the most competitors in this setting as shown in Table \ref{rlmainresults}. In spite of this, IDOL surpassed all the existing strong systems by an obvious margin in nearly every evaluation metric except the accuracy on the LogiQA test set. Apparently, from the results on BERT and ALBERT in Table \ref{rlmainresults} and results on DeBERTa in Table \ref{reclordeberta}, we can see that IDOL has significant advantages over other opponents as well. In summary, IDOL is highly effective in logical reasoning MRC with state-of-the-art performance and this benefit can be generalized to different pre-trained models even to the recent large-scale and strong ones.

\textbf{Prompting} Although the scale of Dev-30 for the pilot test on LLM is small, the results displayed in Table \ref{llmresults} inspired us to some extent. Generally, IDOL is still competitive in the era of LLM. On ReClor, it achieved an accuracy of 80\% while the best result from LLMs is 70\% (ChatGPT with Chain-of-Thought prompting). Even though GLM-130B realizes an accuracy of 50\% on LogiQA in Zero-Shot setting surprisingly (slightly higher than 43.3\% by IDOL), IDOL has an obvious advantage compared with other settings and other LLMs. Additionally, there is an interesting phenomenon that chain-of-thought prompting brings negative effects on LLMs except for ChatGPT on ReClor, which is not consistent with the findings in \citet{cot}.

\begin{table}[t]
\centering
\begin{tabular}{lcccc}
\toprule
 & \multicolumn{4}{c}{\textbf{Setting}} \\
\textbf{Models} & ZS & FS & CoT & FT \\ \midrule

\small\textit{{ReClor}} & & & & \\ 
GPT-3.5 & 56.7 & 50.0 & 46.7 & - \\
ChatGPT  & 63.3 & 63.3 & 70.0 & - \\
GLM-130B & 46.7 & 40.0 & 23.3 & - \\
IDOL & - & - & - & \textbf{80.0} \\  \midrule 

\small\textit{{LogiQA}} & & & & \\ 
GPT-3.5 & 30.0 & 10.0 & 13.3 & - \\
ChatGPT & 33.3 & 36.7 & 23.3 & - \\
GLM-130B & \textbf{50.0} & 36.7 & 26.6 & - \\
IDOL & - & - & - & 43.3 \\ 

\bottomrule
\end{tabular}
\caption{\label{llmresults}Results on ReClor and LogiQA from LLMs and IDOL. ZS: Zero-Shot prompting. FS: Few-Shot prompting. CoT: Chain-of-Thought prompting. FT: Fine-Tuning.} 
\end{table}

\begin{table*}
\small
\centering
\begin{tabular}{m{2cm}<{\centering}m{13cm}}
\toprule[1pt]
\textbf{Setting} & \makecell[c]{\textbf{Template}} \\ \midrule
\textbf{Zero-Shot} & \textit{The passage is \texttt{[PASSAGE]}. The question is \texttt{[QUESTION]}. Here are 4 choices for it and they are \texttt{[CHOICES]}. Which one should I choose? Thanks.} \\ \midrule

\textbf{Few-Shot} & \textit{\texttt{[Example A]} \texttt{[Example B]} \texttt{[Example C]} The passage is \texttt{[PASSAGE]}. The question is \texttt{[QUESTION]}. Here are 4 choices for it and they are \texttt{[CHOICES]}. Which one should I choose? Thanks.} \\ \midrule

\textbf{Chain-of-Thought} & \textit{The passage is \texttt{[PASSAGE]}. The question is \texttt{[QUESTION]}. Here are 4 choices for it and they are \texttt{[CHOICES]}. You can analyze like this, \texttt{[Thought Process]}. So the answer is \texttt{[Answer]}. The passage is \texttt{[PASSAGE]}. The question is \texttt{[QUESTION]}. Here are 4 choices for it and they are \texttt{[CHOICES]}. Which one should I choose? Thanks.} \\ 

\bottomrule[1pt]
\end{tabular}
\caption{Templates and examples for LLM prompting in different settings.}
\label{table:llmtemplates}
\end{table*}

\subsubsection{Other MRC Datasets} \label{othermrc}
For testing whether IDOL could also benefit on types of MRC tasks or maintain the original abilities, we conducted a series of experiments based on RoBERTa as the backbone model. The results are displayed in the middle part of Table \ref{robertageneralizationtable} where we compare the original model, the model further pre-trained with only MLM on LGP and the model further pre-trained with IDOL. We evaluate the models in each setting with 4 different seeds and report the average value. It is apparent that IDOL performs better on both RACE and SQuAD 2.0 in each evaluation metric (although the effects are not as big as those on ReClor or LogiQA), which implies that IDOL indeed helps on general MRC tasks while achieving significant improvement in logical reasoning ability.

\subsubsection{General Understanding Ability}
Following the experiment configuration in section \ref {othermrc}, we planned to find out what kind of effect would IDOL have on other types of natural language understanding tasks which help to reflect the general understanding ability of pre-trained language models. We evaluate the models in each setting with 4 different seeds and report the average value. From the results presented in the right part of Table \ref{robertageneralizationtable}, we can easily find that although IDOL falls behind on MNLI and exceeds the other two competitors on STS-B, the differences in all the evaluation metrics are quite small. Therefore, we could conclude that IDOL retains the general language understanding ability from the original pre-trained model successfully during the process of becoming stronger in logical reasoning.

\section{Ablation Study} 

In this section, we conducted a series of ablation experiments about the multiple logical indicators we used in both fine-tuning and pre-training phases. We evaluate the models based on RoBERTa with 4 different seeds and report the average value.

\subsection{Indicators in Fine-tuning}

As introduced in section \ref{logicindicators}, we defined 5 classes of logical indicators that reflect various logical relations among text logical units and we make use of all of them in IDOL. To figure out whether the 5 types are of equal importance in logical reasoning MRC, we conducted a set of controlled experiments where certain types of indicators are removed from the ReClor train set as the fine-tuning train dataset in each setting. 

From the results displayed in Table \ref{lgidcrmresults}, it is obvious from the last column that logical indicators indeed play an important role in logical reasoning-related text understanding since the loss of all indicators decreases accuracy by 4 to 7 points. In detail, we can conclude that the negative and adversative indicators influence the most by comparing the gaps between pre-training on the original LGP and the dataset without individual types of indicators.

\begin{table}[t]
\small
\centering
\setlength\tabcolsep{4pt} 
\begin{tabular}{lcccccc}
\toprule
\multirow{2}*{\textbf{Models}} & \multicolumn{6}{c}{\textbf{ReClor Train Set}} \\ 
& --- & PMI\&CLI & NTI & ATI & CNI & ALL \\ \midrule
RoBERTa & 62.7 & 64.0 & 59.7 & 61.7 & 63.7 & 59.1 \\
\hspace*{3mm} + MLM & 65.0 & 64.9 & 61.8 & 61.5 & 64.5 & 59.9 \\ 
\hspace*{6mm} + LCP & 66.8 & 63.8 & 62.7 & 63.4 & 64.2 & 60.5 \\ 
\bottomrule
\end{tabular}
\caption{\label{lgidcrmresults} Results of fine-tuning with datasets obtained by removing certain types of logical indicators in the original ReClor train set and testing on the development set. The first row under ``ReClor Train set'' in each column indicates what indicators are removed from LGP. ``---'': the original LGP. ``PMI\&CLI'': both premise and conclusion indicators are removed. ``ALL'': no logical indicators left.}
\end{table}

\subsection{Indicators in Pre-training}


Now that logical indicators have been proven to be effective in fine-tuning stage, we believe they also help with the pre-training stage. Therefore, we arranged a series of experiments on gradually incorporating more logical indicators from not leveraging any indicators (MLM), only making use of PMI and CLI (LCP-2), adding LUI to LCP-2 (LCP-3), to taking advantage of all 6 types of logical indicators (LCP). 

\begin{figure}
\centering
\includegraphics[scale=0.20]{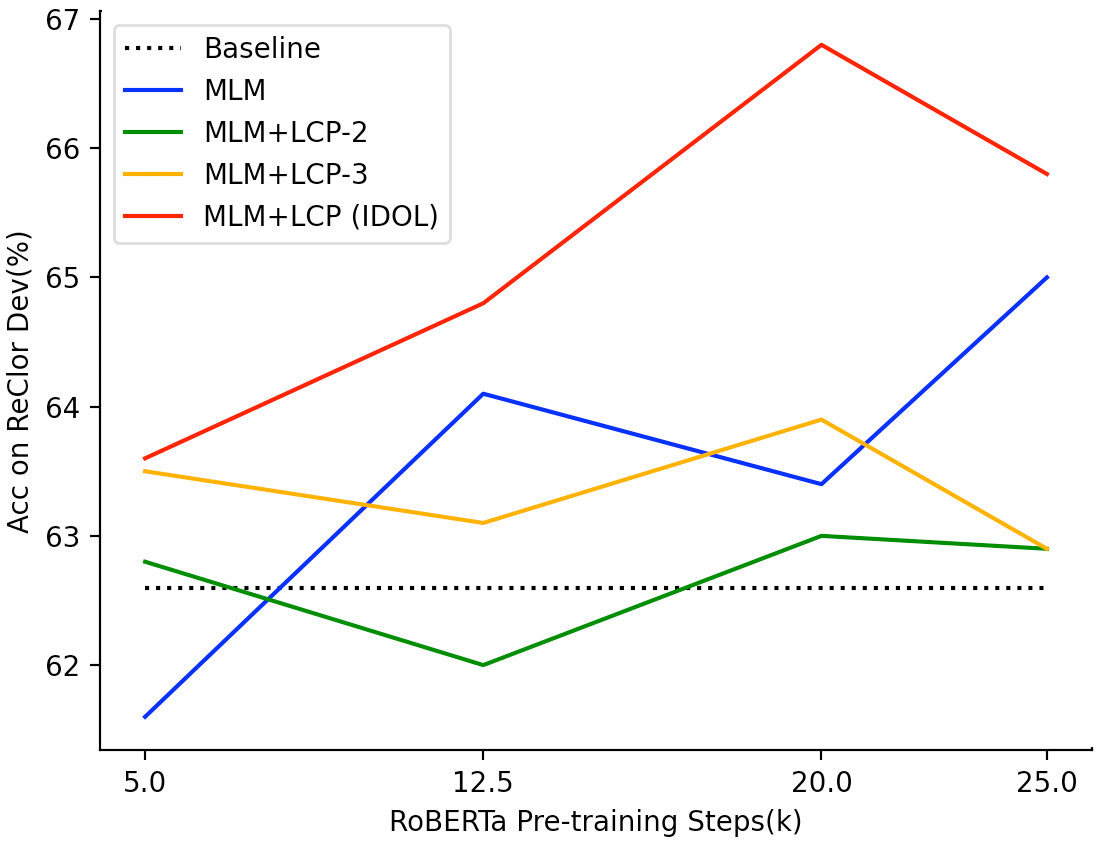}
\caption{The results on ReClor development set of models with different tasks on RoBERTa during the pre-training. LCP-2: LCP only with PMI and CLI. LCP-3: LCP only with PMI, CLI, and LUI. Baseline: fine-tuning with the original RoBERTa.}
\label{fig:diffpretasks}
\end{figure}

From the lines displayed in Figure \ref{fig:diffpretasks}, it is clear that models perform better while leveraging a greater variety of logical indicators since the red line (IDOL) is positioned significantly higher than green and yellow lines representing pre-training tasks that utilize fewer types of logical indicators. According to the results in Table \ref{lgidcrmresults}, PMI and CLI brought the least difference in the model performance on ReClor. The LCP-2 and LCP-3 mainly rely on the two types, and introducing a new special token [LGMASK] inevitably brings noise during model training and further widens the gap between pre-training and down-stream tasks, so that they perform even not better than the original MLM. Additionally, in the aspect of overall trends, the model pre-trained with IDOL is becoming stronger gradually during the process of pre-training, which certifies the effectiveness of our designed task targeted at logical indicators.

\section{Conclusion and Future Work}


In this paper, we proposed an easy-to-understand further pre-training method IDOL which fully exploits the logical information provided by 6 types of logical indicators and is proven effective on different pre-trained language models while keeping them competitive on many other kinds of downstream tasks. Particularly, IDOL achieves state-of-the-art performance on logical reasoning machine reading comprehension tasks.

With respect to future work, we plan to leverage the sentence-level or passage-level logical features in the meantime and integrate it with IDOL to generate a stronger multi-task further pre-training method for improving the logical reasoning ability of pre-trained language models. Moreover, we decide to redesign the IDOL task and find out whether logical indicators also play an important role in those generative pre-trained models as well. Furthermore, we will explore the way of combining IDOL with prompting to find a better method to elicit the reasoning abilities of LLMs.


\section{Limitations}

First of all, IDOL relies on a customized dataset that is filtered out from Wikipedia pages with the help of many pre-defined logical indicators. Inevitably, this will introduce a certain amount of artificial bias. If an automatic method for logical indicator extraction based on something like hidden representations from neural network models is put forward, it would be beneficial to narrow the gap between the dataset preparation and logical pre-training. 

In addition, in the field of pre-training task design, there have been a lot of different but effective approaches proposed. For example, in \citet{pert}, the authors presented a pre-training task named PERT which requires the models to recover the original token sequences under the background of that different token permutation within a certain range would not affect Chinese text understanding. This method only depends on the original texts, but IDOL introduces one more special token, which widens the gap between pre-training and fine-tuning to some extent.


\bibliography{anthology,custom}
\bibliographystyle{acl_natbib}

\appendix





\end{document}